\documentclass[10pt, a4paper]{article}
\usepackage{lrec}
\usepackage{graphicx}
\usepackage{tabularx}
\usepackage{soul}
\usepackage{times}
\usepackage{latexsym}
\usepackage{url}
\usepackage{amsmath}
\usepackage{graphicx}
\usepackage{subfig}
\usepackage{float}
\usepackage{placeins}
\usepackage{xcolor}
\usepackage{epstopdf}
\usepackage[latin1]{inputenc}
\usepackage{titlesec}
\titleformat{\section}{\normalfont\large\bf\center}{\thesection.}{1em}{}
\titleformat{\subsection}{\normalfont\SmallTitleFont\bf\raggedright}{\thesubsection.}{1em}{}
\titleformat{\subsubsection}{\normalfont\normalsize\bf\raggedright}{\thesubsubsection.}{1em}{}
\renewcommand\thesection{\arabic{section}}
\renewcommand\thesubsection{\thesection.\arabic{subsection}}
\renewcommand\thesubsubsection{\thesubsection.\arabic{subsubsection}}
\usepackage{hyperref}
\usepackage{xstring}

\newcommand{\carlos}[1]{\textcolor{black}{#1}}
\newcommand{\markda}[1]{\textcolor{black}{#1}}
\newcommand{\john}[1]{\textcolor{black}{#1}}
\newcommand{\carlosII}[1]{\textcolor{black}{#1}}

\newcommand{\markdaIII}[1]{\textcolor{black}{#1}}
\newcommand{\carlosIII}[1]{\textcolor{black}{#1}}

\title{Inherent Dependency Displacement Bias of Transition-Based Algorithms}

\name{Mark Anderson\qquad Carlos G\'{o}mez-Rodr\'{i}guez}

\address{Universidade da Coru\~na, CITIC \\
  FASTPARSE Lab, LyS Research Group, Departamento de Computaci\'on y Tecnolog\'ias de la Informaci\'on\\
         Campus Elvi\~{n}a, s/n, 15071\\ 
  A Coru\~{n}a, Spain\\{\tt \{m.anderson,carlos.gomez\}@udc.es}}

\abstract{
A wide variety of transition-based algorithms are currently used for dependency parsers. Empirical studies have shown that performance varies across different treebanks in such a way that one algorithm outperforms another on one treebank and the reverse is true for a different treebank. There is often no discernible reason for what causes one algorithm to be more suitable for a certain treebank and less so for another. In this paper we shed some light on this by introducing the concept of an algorithm's inherent dependency displacement distribution. This  characterises the bias of the algorithm in terms of dependency displacement, which quantify both distance and direction of syntactic relations. We show that the similarity of an algorithm's inherent distribution to a treebank's displacement distribution is clearly correlated to the algorithm's parsing performance on that treebank, specifically with highly significant and substantial correlations for the predominant sentence lengths in Universal Dependency treebanks. We also obtain results which show a more discrete analysis of dependency displacement does not result in any meaningful correlations.
\\ \newline  }

\begin{document}

\maketitleabstract
\section{Introduction}
Dependency parsing, and in particular the transition-based family of parsers, has a large variety of parsing algorithms to choose from. When comparing different algorithms, empirical results on collections of corpora often show differences in accuracy that can heavily vary across different languages or treebanks, so that a given algorithm can be the best choice for one corpus while being outperformed in another.

However, despite these nontrivial patterns in accuracy variations having been observed in many experiments in the last decade, both with
non-neural and neural implementations \carlos{\cite{nivre2008,ballesteros2013,chen2014,fernandez2016}}, very little is known about what makes an algorithm more fitting for a corpus beyond obvious facts (like non-projective algorithms being better for highly non-projective treebanks). This makes the practical choice of a particular algorithm for a language a matter of trial and error. For example, Malt\-Optimizer chooses between projective and non-projective algorithm according to the amount of non-projective dependencies observed in the treebank, but then the choice of a specific algorithm among projective (or non-projective) options is made by running all of the projective (or the non-projective) algorithms and choosing the one that achieves the highest accuracy \cite{ballesteros2012}.

\paragraph{Contribution}
We shed some light on this issue by performing two complementary analyses:
\begin{enumerate}
    \item An analysis of the performance of transition-based algorithms with respect to dependency displacement (signed distance) which highlights that a shallow level of analysis is not sufficient to account for the aforementioned differences.
    \item An analysis focusing on the distribution of dependency displacements which shows the performance of a transition-based algorithm on a given corpus can be partly explained by the similarity of the dependency displacement distribution of the corpus \carlos{(sampled from the test set)} and what we call the algorithm's \emph{inherent distribution}. \carlos{This is a random distribution that describes the bias of the algorithm in terms of dependency lengths and directions, and it can be approximated} by running the algorithm in random mode. This entails assigning a uniform probability to each of the transitions available for a given configuration.
\end{enumerate}
Experiments with Universal Dependencies show a highly significant correlation between said similarity and the accuracy of algorithms on different treebanks for the most prevalent sentence lengths in the corpora.

\section{Related work}

\subsection{How language-specific performance differs across algorithms}

\carlos{With the proliferation of dependency treebanks for multiple languages, various papers performed comparisons including multiple transition-based parsing algorithms, where language-specific differences in performance across algorithms are apparent.}

\carlos{One of the first papers that included a large number of languages and algorithms is that of \newcite{nivre2008}, who}
found language-specific differences between the performance of Arc-Standard and Arc-Eager. \carlos{While \markda{they} hypothesized that the proportion of left arcs in a language could \markdaIII{account for }
these differences, the evidence was not conclusive.} 

Other analyses have found differences in performances between Arc-Eager and Arc-Standard such as \newcite{ballesteros2013} or \newcite{fernandez2016}\carlos{, but they provided no explanation for this phenomenon.} \newcite{ballesteros2013} did consider the effect the dummy root placement has on different algorithms. They found that the placement is not trivial and had a noticeable effect on the performance of Arc-Eager. 




The differences between algorithms have also been observed with different architectural implementations, namely neural networks. \newcite{chen2014} found that for some treebanks Arc-Eager performed better whereas for others Arc-Standard performed better.

\subsection{Analysis of parsing errors}

\carlos{There 
\markda{have been analyses on} the strengths and weaknesses of different parsers by comparing the errors they make on different kinds of dependencies, with dependency distance and direction often being considered.}

The most comprehensive analysis on the performance of dependency parsers came from \newcite{mcdonald2011}, who compared \carlos{a graph-based parser with a transition-based parser}. They investigated \carlos{the relative strengths of each of the parsing paradigms with respect to factors like dependency length, distance to root, and sentence length. They found 
that the transition-based parser performed worse for longer dependencies and those closer to the root, due to error propagation, and better in the converse cases. However, while this analytical methodology gave interesting insights when comparing parsers from very different paradigms, it is not fine-grained enough to draw conclusions when the comparison is 
\markda{between} transition-based algorithms, as we will verify in Section \ref{sec:dispanalysis}}.

\newcite{de2017} investigated the differences in performances \carlos{when using the same transition-based algorithm} with a neural network implementation, namely UDPipe  \cite{straka2017tokenizing}, and with a classical implementation, MaltParser \cite{nivre2007maltparser}. 
They observed a similar trend for both implementations with respect to the change in F1 score for different dependency distance bins. 

\john{More recently, \newcite{kulmizev-etal-2019-deep} evaluated the difference in performance between graph-based and transition-based parsers and observed that the use of contextualised word embedding off-set the typical error-propagation found when using transition-based parsers.}

\newcite{rehbein2017} investigated what made certain corpora harder to parse than others by looking at how 
\markda{the dependency encoding scheme} affects dependency length and arc direction entropy. They, however, did not observe a correlation in parsing accuracy for different dependency lengths resulting from different encoding styles, but they did find a correlation 
\carlos{with respect to}
arc direction entropy. This is somewhat corroborated by a previous analysis undertaken by \newcite{gulordava2016}, who modified treebanks to minimise dependency length and the arc direction entropy and \markda{obtained an increase in parser performance using these treebanks.}

Additionally, \newcite{kirilin2015} evaluated error patterns (over- and underproduction of certain structures); \newcite{goldberg2010} 
\carlosII{made a similar analysis}
by classifying parser output and test sentences based on underproduced \markda{(indicating test) and overproduced (indicating prediction)} structures; \newcite{dickinson2010} also showed that algorithms generate idiosyncratic structures not found in the training data; and \newcite{schwartz2012} found that certain structures work better with certain heads.
Together these findings suggest that algorithms have some sort of bias towards generating certain types of structures.

\subsection{Dependency distance} There has also been more linguistic-specific work undertaken with respect to dependency distance\carlos{, which can be relevant for parsing}. \markda{For example, \newcite{gibson2000dependency} introduced \emph{dependency locality theory} (DLT) which postulates that dependency distances are minimised in order to make parsing linguistic input more cognitively friendly. \newcite{liu2017} tested this theory by evaluating} how syntactic patterns are governed by  dependency distance. They \carlos{provide\markda{d} a review of} 
psycholinguistic experiments and treebank analyses that \markda{suggest} natural languages have a tendency to minimize dependency distance, arguing that this makes sense under 
working-memory restrictions. So if longer distance dependencies are difficult for humans to parse, it would make sense then that parsing algorithms would struggle in a similar vein. 

\markda{Many analyses have
\markda{observed this} tendency to minimise dependency distance, \markda{further corroborating} this hypothesis} \cite{cancho2004,liu2008,liu2007,buch2006,futrell2015,temperley2018}. \markda{However, the  extent to which this restriction is adhered has been observed to vary significantly across languages \cite{jiang2015,gildea2010}}. 
\markda{Other works have investigated different syntactic traits of languages associated with dependency length such as highlighting a correlation with an increase in dependency length and free-order languages \cite{gulordava2015} and with an increase in non-projective dependencies \cite{ferrer2016,gomez2017}.}

In response 
to \newcite{liu2017}, \carlos{\newcite{GomPoLR2017} hypothesised that the good practical results achieved by transition-based algorithms 
\markda{could be because} they are biased towards short dependencies. \markda{This is 
substantiated by
\newcite{eisner2010} 
who improved parser performance by imposing upper bounds on dependency length and using dependency lengths as a parsing feature.} In another response to \newcite{liu2017},} \newcite{hudson2017} highlighted mean dependency distance varies significantly between treebanks 
and that the direction of dependencies 
could impact parsing difficulty. \newcite{liu2010}  and \newcite{jiang2015} actually found that dependency direction analysis 
can be used to typologically classify 
language\markda{s}. Hence, for our analyses we use dependency displacement which quantifies both length and direction as defined in Equation \ref{eqn:displacement}.

\section{Data and setup}


\carlos{We report two levels of analysis to explain performance differences between different transition-based algorithms. The first is based on measuring the differences in performance for each dependency displacement. It transpires that this does not offer any explanation how language-specific performance differs between algorithms. For this, we need the second analysis, based on the concept of an inherent displacement distribution of each algorithm.}


We evaluate \carlos{3 projective algorithms} (Arc-Standard, Arc-Eager and Covington) and \carlos{2 unrestricted non-projective algorithms} (Covington and Swap-Eager). The results for each algorithm were obtained by running Maltparser v1.91, which has the benefit of having multiple algorithms implemented of which several are projective and several are non-projective. \carlos{While there are more modern systems that outperform MaltParser in 
accuracy, none of them provide such a range of algorithms.\footnote{For example, UDPipe has one projective, one partially non-projective and one unrestricted non-projective parser.}} Furthermore, as we are focused on the algorithms, the architecture of the implementation is \markdaIII{considered a controlled variable.} 
Also, \newcite{de2017} showed that the change in accuracy with respect to dependency length follows a similar trend for MaltParser and UDPipe (a neural network implementation)
, so it is justifiable to use MaltParser for analyses on algorithms.

\john{Furthermore, MaltParser is potentially better suited for this analysis as we want to evaluate the performance of algorithms against one another and using a more robust system that can more readily overcome the inherent biases we show in this paper would actually obscure this. For this reason too, we do not finetune the feature functions for each language as it is obvious that different features will benefit different languages more or less depending on the linguistic features of a given language. So these features are kept constant for each treebank and should be considered controlled variables for the following analyses.}

\john{Beyond these experimental considerations, MaltParser is still competitive with respect to parsing accuracy even if it is not state-of-the-art and it is very efficient which makes it much more readily deployable when compared to large and unwieldy neural networks.}

Version 2.1 of the Universal Dependency treebanks was used for all of the subsequent analysis \cite{ud21}.

\section{Dependency displacement}
\label{sec:dispanalysis}

\subsection{Analysis details}
Similar to the analysis undertaken by \cite{mcdonald2011}, we investigated how parsing performance varies based on the dependency displacement. Whereas they compared a graph-based and a transition-based parser, we have compared different transition-based parsing algorithms. Also, we look at the effect dependency displacement, $s_\textrm{{dep}}$, has on the performance of the algorithms, rather than dependency distance.
\carlos{Dependency displacement is defined as:}
\begin{equation}
    s_{\textrm{dep}} = \textrm{x}_{\textrm{head}} - \textrm{x}_{\textrm{dependent}} 
    \label{eqn:displacement}
\end{equation}
where $x_{i}$ refers to the position of word \emph{i} in a given sentence. \markda{So a right arc of length 3 would have a displacement of -3 and conversely a left arc of length 3 would have a displacement of 3.} \newcite{hudson2017} highlighted that different languages have a tendency towards being head-final (OSV or SOV), head-medial (OVS or SVO), or head-initial (VOS or VSO) and argue that the direction of dependencies cannot be ignored when analysing 
dependency distances. \carlos{In addition, dependency direction was hypothesised by \newcite{nivre2008} to 
\markda{affect how} language-specific performance differs between parsers.}

We evaluate the attachment precision and recall in order to have a more fine-grained analysis similar to \newcite{de2017}.


For this analysis we used all 76 treebanks that contained a training and test set.
\subsection{Results}
\begin{figure}
  \centering
    \captionsetup[subfigure]{oneside,margin={0cm,0cm}}
  \subfloat[\label{fig:precision}]{\includegraphics[width=0.99\linewidth]{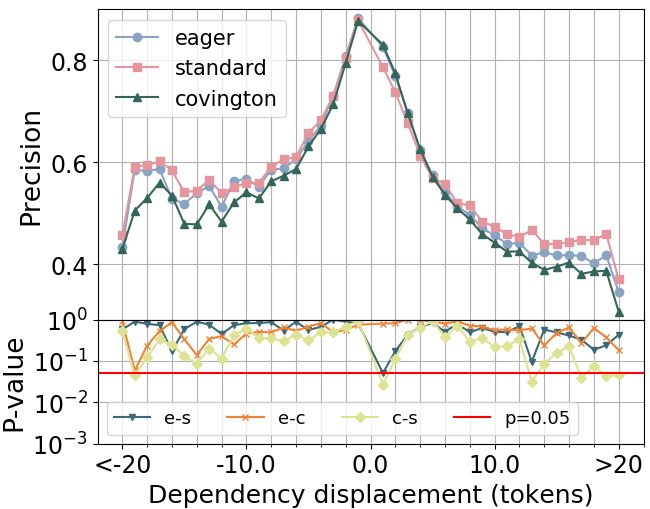}}
  \hspace{0.5cm}
  \subfloat[\label{fig:recall}]{\includegraphics[width=0.99\linewidth]{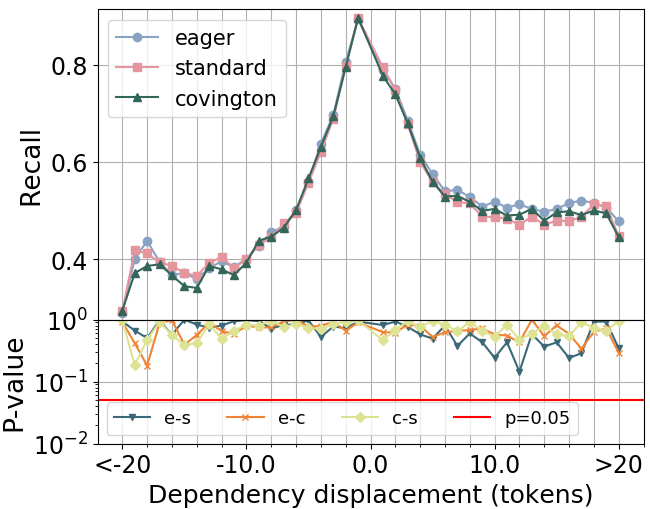}}
    \caption{Attachment precision (a) and recall (b) for the three projective algorithms used: Arc-Eager (eager, blue, circle), Arc-Standard (standard, magenta, square), Covington (covington, green, triangle). The corresponding p-values (derived from a t-test respectively using the average precision and recall and the corresponding standard deviation across treebanks for each displacement value for each algorithm) are shown below: Arc-Eager and Arc-Standard (e-s); Arc-Eager and Covington (e-c); and Covington and Arc-Standard (c-s). No statistically significant differences are observed for any comparison with regard to attachment precision or recall.}
  \label{fig:precisionRecall}
\end{figure}

Figure \ref{fig:precisionRecall} shows that this coarse analysis does not capture any statistically meaningful variation across projective algorithms with regard to attachment precision or recall. It is interesting to note that long-distant displacement to the right and to the left follow different trends for both precision and recall, with precision being higher for right arcs (negative displacement) but recall being higher for left arcs (positive displacement).

\begin{figure}
  \centering
    \captionsetup[subfigure]{oneside,margin={0cm,0cm}}
  \subfloat[\label{fig:precison_np}]{\includegraphics[width=0.99\linewidth]{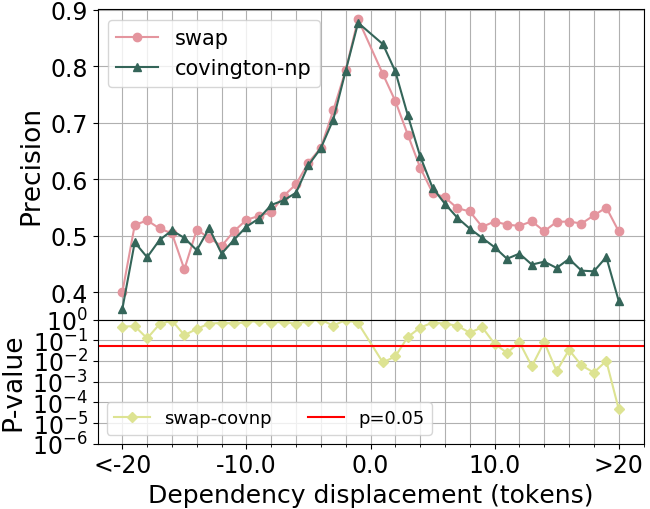}}
  \hspace{0.5cm}
  \subfloat[\label{fig:recall_np}]{\includegraphics[width=0.99\linewidth]{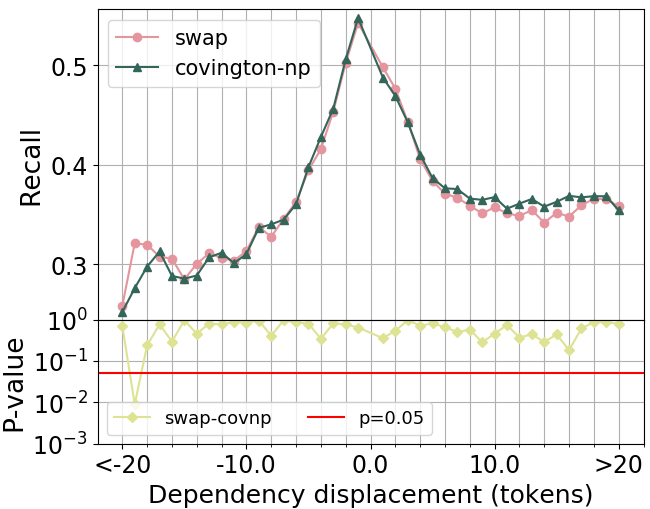}}
  \caption{Attachment precision (a) and recall (b) for the two non-projective algorithms tested: Swap-Eager (swap, magenta, circle) and non-projective Covington (covington-np, green, triangle). The corresponding p-values (derived from a t-test respectively using the average precision and recall and the corresponding standard deviation across treebanks for each displacement value for each algorithm) are shown below: Swap-Eager and non-projective Covington (swap-covnp). Almost no statistically significant differences are observed for any comparison with regard to attachment precision or recall, except for left arcs (positive displacement) with regard to precision.}
  \label{fig:precisionRecallNP}
\end{figure}

Figure \ref{fig:precisionRecallNP} shows the corresponding results for the non-projective algorithms. 
\markda{W}e see in Figure \ref{fig:precison_np} the only statistically meaningful difference between algorithms.  Precision for left arc dependencies is higher for Swap-Eager than non-projective Covington. It is also of note that the non-projective precision results are much more symmetric with regard to dependency displacement 
than those of
the projective algorithms. The\markdaIII{se} results 
show that \markdaIII{there is a need for a more fine-grained analysis.}

\section{Comparing displacement distributions}
\subsection{Analysis overview}
\begin{figure}[h]
    \centering
        \includegraphics[width=0.99\linewidth]{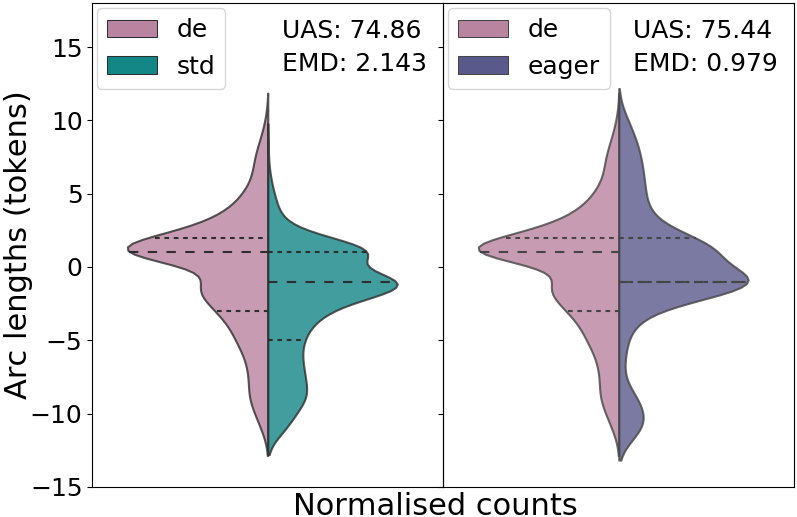}
    \caption{Example comparisons between the dependent displacement distribution of the German-GSD treebank (de, magenta) for sentences of length 10 to 12 in version 2.1 of the Universal Dependency treebanks and the inherent displacement distributions of two algorithms: Arc-Standard (std, green) and Arc-Eager (eager, purple). The corresponding UAS and EMD values are displayed.}
    \label{fig:distros}
\end{figure}

In this analysis we test the hypothesis that the similarity of the dependency displacement distribution generated by the latent biases of an algorithm, as defined as its inherent displacement distribution in Section \ref{section:latent}, and the actual distribution of a treebank can account for the difference in performance 
\carlos{across} 
parsing algorithms. We measure the difference between these distributions by using the Wasserstein distance - also known as the earth mover's distance (EMD). \markda{It} can be intuitively thought of as the cost or work required to change one distribution into another \carlos{by moving distribution mass} \cite{rubner98}. An example comparison for the German test data for sentences of length 10 to 12 tokens is shown in Figure \ref{fig:distros}, where Arc-Standard is seen to perform worse than Arc-Eager by 0.58 UAS and correspondingly has a higher EMD. We evaluate dependency displacements according to sentence-length bins, as it has been shown that dependency 
\carlos{distance (and thus displacement) distributions}
change with sentence length, both in real sentences and in random models \cite{ferrer2014}.

\subsection{Inherent displacement distributions}\label{section:latent}
Let $P = (C,T,c_s,C_t)$ be a transition-based algorithm where $C$ is the set of possible configurations, $T$ the set of transitions, $c_s$ an initialization function mapping a sentence length $k$ to an initial configuration $c_s(k)$, and $C_t$ a set of terminal configurations. We assume configurations in $C$ to be of the form $(D,A)$, where $A$ is a set of dependency arcs built so far, and $D$ is a state of the data structures associated with the algorithm (e.g. a stack and a buffer, for stack-based algorithms).

Then, we define the \emph{inherent distribution} of $P$ for sentences of length $k$, written $\iota_k(P)$, as the discrete probability distribution of the random variable generated by the following process:\begin{itemize}
   \item Start at the initial configuration $c_s(k)$.
   \item At each configuration, let $t_1 \ldots t_q$ be the available transitions. Choose one of them randomly with probability $1/q$, and go to the 
   \carlosII{resulting next configuration.}
   \item The process ends when a terminal configuration $c_t = (D_t, A_t) \carlosII{\in C_t}$ has been reached. Then, choose a dependency arc \carlos{uniformly} at random from $A_t$ and take its displacement \carlos{as the value of the random variable}.
\end{itemize}Note that the inherent distribution of an algorithm does not depend on the contents of the particular sentence being parsed in the stochastic process, but just its length. Therefore, it can be seen as a variable that describes the distribution of displacements that the algorithm is ``biased'' to produce, in the absence of any training data. \john{The transition is selected using a uniform probability across all possible transitions for a given configuration as there is no underlying reason why an algorithm would select one transition over another \emph{without} using the feature function at a given timestep.} Our hypothesis is that a given language or corpus will be parsed more accurately by algorithms whose inherent distribution is closer \carlos{(as measured by the EMD)} to the actual observed displacements in that language or corpus. 

While the inherent distribution of an algorithm for sentences of length $k$ would be difficult to obtain analytically, \john{especially for the algorithms that support arbitrary non-projectivity where exact inference is intractable \cite{mcdonald2007complexity},}
in practice we will approximate it by running a number of simulations of the above stochastic process.

The above definition can be extended to corpora (or subsets of them\carlos{, such as the sentence-length bins we use in this paper}). Let $S$ be a set of $n$ sentences containing $n_k$ sentences of length $k$, for a range of values of $k$. Then, the inherent distribution of $P$ \carlos{with respect to} $S$ is the discrete probability distribution of the random variable generated by taking a random sentence length from $S$ (where each length $k$ is taken with probability $n_k/n$), and then taking a random displacement using the process above. The inherent distribution of $P$ with respect to a set $S$%
\carlosII{, denoted $\iota_S(P)$,}
can be approximated by running a number of simulations of the above stochastic process on all the sentences of $S$.

\subsubsection*{Approximating inherent distributions}
In order to \carlos{approximate} the biased distributions for each parsing algorithm \carlosII{$P$}, we implemented \carlos{a version of each of} them so that they randomly select a transition from the set of available transitions for any given configuration. For each tree in a treebank that fell within the range of a sentence-length bin, a random tree was generated this way. This was done so as to ensure the EMD of different distributions was due to differences in the dependency displacement and to minimize other factors affecting the EMD. \john{In other words, we wanted to obtain inherent distributions that echo the output of a parsing algorithm if it had been run normally (i.e. making predictions based on a feature function).} For each treebank \carlosII{and each sentence length bin $B$ in its test set}, a random displacement distribution was generated 10 times \carlosII{to approximate $\iota_B(P)$}, 
\carlosII{and their average EMD of the observed distribution of displacements in the trees of $B$ was taken.} \john{We opted to run it 10 times individually rather than run it once with more data points as this way we can more easily evaluate the uncertainty of a given inherent distribution. The standard error for the average EMD can be observed in Figures \ref{fig:dUASvsEMD-proj} and \ref{fig:dUASvsEMD-np}. It is clear that the variation across each generated distribution is quite small and that 10 instances are enough to minimise the uncertainty of this procedure.}

\begin{figure}[h]
    \centering  
    \includegraphics[width=0.99\linewidth]{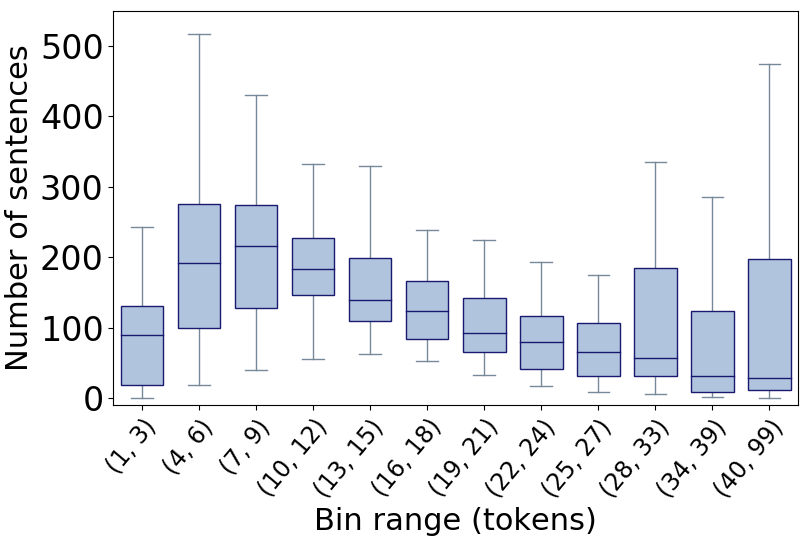}
    \caption{Corresponding stats for each bin used in the subsequent analysis with the average across the 26 treebanks shown and the first and third quartile limits.}   
    \label{fig:binStats}
\end{figure}

 \john{We split the data according to sentence length to account for the differences in arc lengths that arise from longer or shorter sentences. Optimally, we would have undertaken our analyses according to sentence length and would not have used bins, however, the statistics were too low for many treebanks in order to this. The sentence lengths bins used and 
\carlosII{their statistics}
can be seen in Figure \ref{fig:binStats}.}

Because of the varying difficulty of the datasets in the universal dependency collection, we have opted to compare the average EMD against $\delta$UAS, defined as the difference between the performance of an algorithm on a treebank minus the average score \carlos{across algorithms} for that treebank:
\begin{equation}
  \delta \textrm{UAS}_{L_{A}} = \textrm{UAS}_{L_{A}} - \overline{\textrm{UAS}}_{L}
  \label{eqn:deltaUAS}
\end{equation}
where A is the algorithm and L is the treebank.

We then compare the $\delta$UAS for an algorithm against its average EMD. We do this for 26 treebanks from version 2.1 of the Universal Dependencies treebanks. These languages were selected based on their size. We removed all languages with less than 1,000 trees in the training set and in the test set. This was necessary because if a treebank was too small then most sentence-length bins would not have enough stats to compute a meaningful EMD. 

We split our analysis into projective and non-projective algorithms. The performance between projective and non-projective algorithms on certain datasets would be dominated by the percentage of non-projective arcs in the data and would potentially cloud any effect that the displacement distribution similarities might have. \newcite{nivre2008} found a strong correlation between the percentage of non-projective dependencies and the improvement in accuracy for a non-projective parser (r=0.815, p=7.0x10$^{-4}$) using a much more limited dataset (CoNLL-X shared task 2006). Furthermore, the search space affects the random distributions (for example, non-projective trees in general have larger average dependency lengths than projective trees\john{, see for example \cite{ferrer2016}}) and this could also be a confounding factor. So by separating projective and non-projective algorithms, we have made the search space a fixed factor.

Finally, we also compared algorithms directly. We did this by comparing $\Delta$UAS and $\Delta$EMD, defined as:
\begin{align}
    \Delta\textrm{UAS}_{L} &= \textrm{UAS}_{L_{A_{1}}} - \textrm{UAS}_{L_{A_{2}}} \label{eqn:DeltaUAS}\\
    \Delta\textrm{EMD}_{L} &= \overline{\textrm{EMD}}_{L_{A_{1}}} - \overline{\textrm{EMD}}_{L_{A_{2}}} \label{eqn:DeltaEMD}
\end{align}
where $A_1$ is the first algorithm, $A_2$ is the second, and $L$ is the treebank.
\subsection{Results}
 There is a lack of meaningful correlation for both projective and non-projective when looking at the displacement distributions for unbinned treebanks. The correlation and corresponding p-value for the projective algorithms were r = -0.045 and p = 6.97x10$^{-1}$. For the non-projective algorithms they were r = -0.252 and p = 7.17x10$^{-2}$. Neither result is statistically significant nor does either show any strong correlation despite that. This corroborates the findings of \newcite{ferrer2014} and further justifies our binning procedure.

\begin{figure}
    \centering  
    \includegraphics[width=0.99\linewidth]{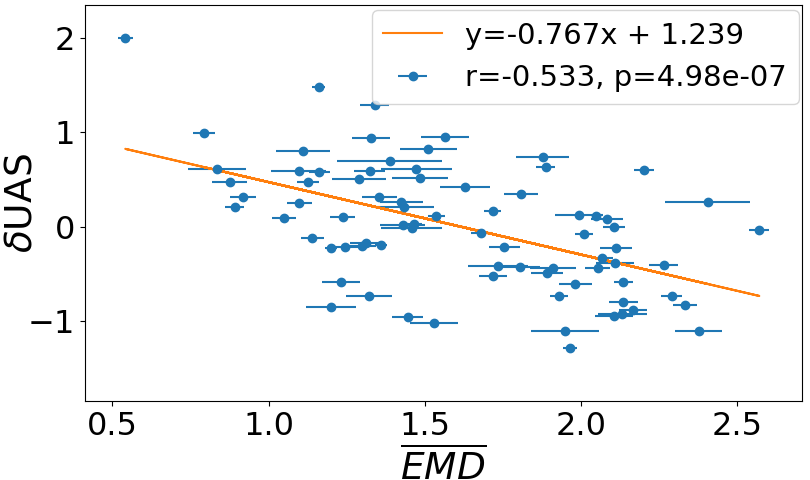}
    \caption{$\delta$UAS (as defined in Equation \ref{eqn:deltaUAS}) for each algorithm against the corresponding average EMD (k=10) for projective algorithms in the 10-12 token sentence-length bin.}   
    \label{fig:dUASvsEMD-proj}
\end{figure}

Figure \ref{fig:dUASvsEMD-proj} shows an example plot comparing $\delta $UAS against $\overline{\textrm{EMD}}$ for the projective algorithms. For this bin (10-12 tokens) there is a strong negative correlation of -0.533. Hence, r$^2$ is 0.284, meaning that the $\overline{\textrm{EMD}}$ accounts for 28.4\% of the variance seen in $\delta $UAS. The correlation is statistically significant \carlos{ ($p=4.98 \times 10^{-7}$)}.

The corresponding results for all of the sentence-length bins can be observed in Table \ref{metrics-table}, where unsurprisingly the shortest sentences \carlos{(lengths 1 to 3)} show no correlation (they are too short for any meaningful difference between dependency displacement distributions) and the correlations start to diminish as the sentences get larger, but 
\carlos{are still statistically meaningful}
until sentence lengths of 25. 
\begin{table}[h]
\begin{center}
\begin{tabular}{cccc}
\textbf{token bin} & \textbf{r} & \textbf{r$^2$} & \textbf{p}  \\ \hline
1-3 & -0.060 & 0.004 & 6.09$\times 10^{-01}$ \\ 
4-6 & -0.401 & 0.161 & 2.75$\times 10^{-04}$ \\ 
7-9 & -0.503 & 0.253 & 2.74$\times 10^{-06}$ \\ 
10-12 & -0.533 & 0.284 & 4.98$\times 10^{-07}$ \\ 
13-15 & -0.526 & 0.277 & 7.49$\times 10^{-07}$ \\ 
16-18 & -0.514 & 0.264 & 1.47$\times 10^{-06}$ \\ 
19-21 & -0.402 & 0.161 & 2.68$\times 10^{-04}$ \\ 
22-24 & -0.304 & 0.093 & 6.78$\times 10^{-03}$ \\ 
25-27 & -0.202 & 0.041 & 7.65$\times 10^{-02}$ \\ 
28-33 & -0.072 & 0.005 & 5.29$\times 10^{-01}$ \\ 
34-39 & -0.034 & 0.001 & 7.70$\times 10^{-01}$ \\
40-99 & 0.139 & 0.019 & 2.43$\times 10^{-01}$ \\ 
\end{tabular}
\end{center}
\caption{\label{metrics-table} Full results for each sentence-length bin for projective algorithms where token bin is the sentence length range, r is the Pearson coefficient of the correlation between $\delta$UAS and the EMD of each algorithm (e.g. as shown in Figure \ref{fig:dUASvsEMD-proj}), $r^2$ is the squared Pearson coefficient which gives an indication of how much variation in the data this correlation accounts for, and finally p is the p-value for a given correlation.}
\end{table}

\begin{figure}
    \centering  
    \includegraphics[width=0.99\linewidth]{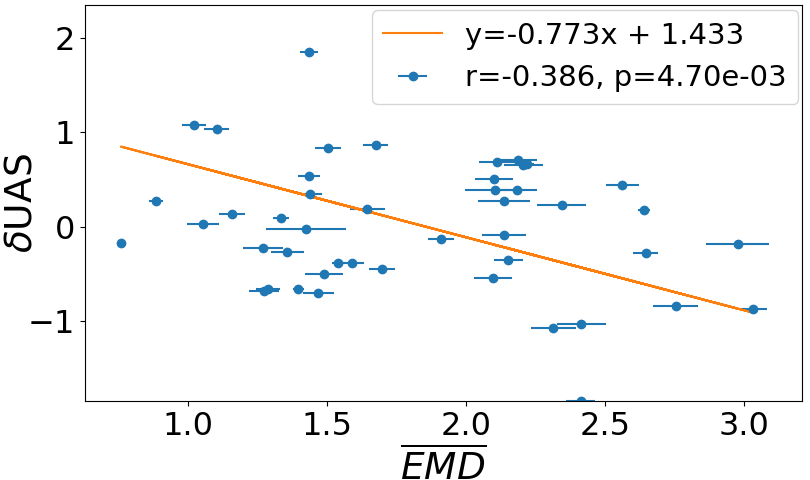}
    \caption{$\delta$UAS (as defined in Equation \ref{eqn:deltaUAS}) for each algorithm against the corresponding average EMD (k=10) for non-projective algorithms in the 10-12 token sentence-length bin.}   
    \label{fig:dUASvsEMD-np}
\end{figure}

Figure \ref{fig:dUASvsEMD-np} shows the correlation for the non-projective algorithms for the same bin as Figure \ref{fig:dUASvsEMD-proj}. It is clear that the correlation is not as strong for the non-projective algorithms, but it is still large enough to be meaningful and is statistically significant. Table \ref{metrics-table-np} shows the results for all of the bins used. Interestingly, the correlation does not diminish so severely for the non-projective algorithms as the sentence length increases as is the case for the projective algorithms. 
\begin{table}[h]
\begin{center}
\begin{tabular}{cccc}
\textbf{token bin} & \textbf{r} & \textbf{r$^2$} & \textbf{p}  \\ \hline
1-3 & 0.001 & 0.000 & 9.92$\times 10^{-01}$ \\ 
4-6 & -0.243 & 0.059 & 8.32$\times 10^{-02}$ \\ 
7-9 & -0.327 & 0.107 & 1.79$\times 10^{-02}$ \\ 
10-12 & -0.386 & 0.149 & 4.70$\times 10^{-03}$ \\ 
13-15 & -0.344 & 0.118 & 1.25$\times 10^{-02}$ \\ 
16-18 & -0.364 & 0.133 & 7.90$\times 10^{-03}$ \\ 
19-21 & -0.344 & 0.118 & 1.26$\times 10^{-02}$ \\ 
22-24 & -0.350 & 0.122 & 1.11$\times 10^{-02}$ \\ 
25-27 & -0.349 & 0.122 & 1.11$\times 10^{-02}$ \\ 
28-33 & -0.347 & 0.121 & 1.17$\times 10^{-02}$ \\ 
34-39 & -0.354 & 0.125 & 1.01$\times 10^{-02}$ \\ 
40-99 & -0.298 & 0.089 & 3.97$\times 10^{-02}$ \\ 
\end{tabular}
\end{center}
\caption{\label{metrics-table-np} Full results for each sentence-length bin for non-projective algorirthms  where token bin is the sentence length range, r is the Pearson coefficient of the correlation between $\delta$UAS and the EMD of each algorithm (e.g. as shown in Figure \ref{fig:dUASvsEMD-np}), $r^2$ is the squared Pearson coefficient which gives an indication of how much variation in the data this correlation accounts for, and finally p is the p-value for a given correlation.}
\end{table}

\begin{figure}
    \centering  
    \includegraphics[width=0.99\linewidth]{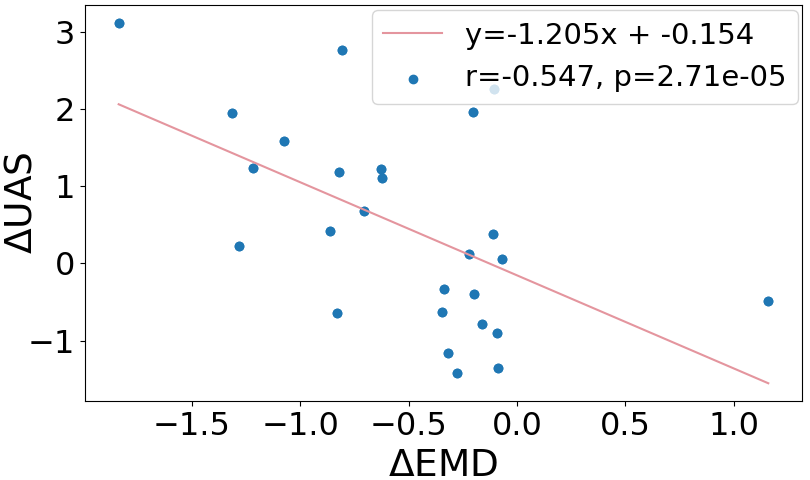}
    \caption{$\Delta$UAS as defined in Equation \ref{eqn:DeltaUAS} against $\Delta$EMD as defined in Equation \ref{eqn:DeltaEMD} comparing Arc-Eager and Arc-Standard (10-12 token sentence-length bin).}   
    \label{fig:EagervsStandard}
\end{figure}

\begin{figure}
    \centering  
    \includegraphics[width=0.99\linewidth]{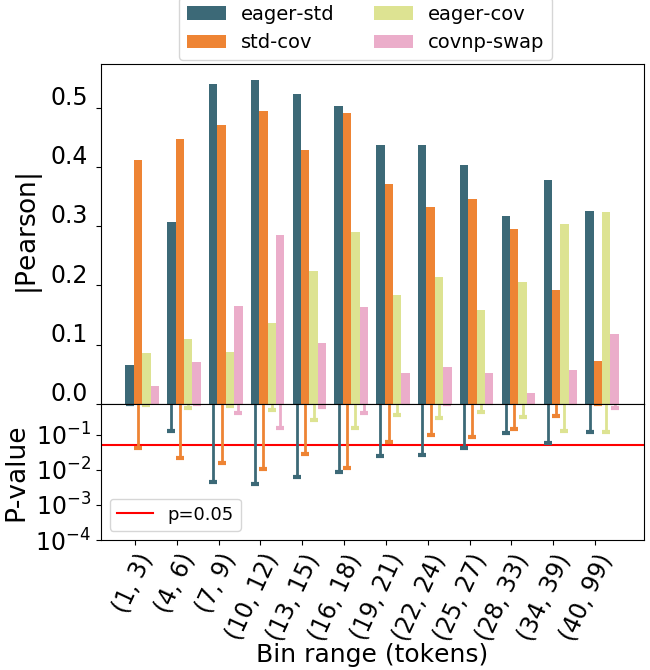}
    \caption{Absolute Pearson coefficients and the corresponding p-values from comparisons between each \carlos{pair of} projective algorithms and the two non-projective algorithms: Arc-Eager and Arc-Standard (eager-std, blue); Arc-Eager and Covington (eager-cov, yellow); Arc-Standard and Covington (std-cov, orange); and non-projective Covington and Swap-Eager (covnp-swap, magenta). Statistically relevant comparisons can be seen between Arc-Standard and the other projective algorithms for mid-range tree lengths.}   
    \label{fig:comparisons-algos}
\end{figure}

Finally, \carlos{focusing on direct comparisons between two algorithms,} Figure \ref{fig:EagervsStandard} shows the comparison between Arc-Standard and Arc-Eager for the same bin as above. A strong negative correlation can be seen which is statistically significant. The results for the direct comparison between the three projective algorithms and the two non-projective algorithms can be seen in Figure \ref{fig:comparisons-algos}. These comparisons yield significant results for Arc-Standard when compared with Arc-Eager and Covington for moderate lengthed sentences (4-25 tokens), but not for the non-projective algorithms and for Arc-Eager and Covington. This suggests the previous analysis of comparing all projective and the two non-projective algorithms is a statistically more powerful means of analysing the effect of the similarity of dependency displacement distribution on algorithm performance.
\section{Discussion}
A coarse analysis focusing on dependency displacements individually does not show a significant difference in performance across algorithms except with regards to the attachment precision for non-projective algorithms. However, the difference 
between an algorithm's biased latent dependency displacement distribution and the target treebank being parsed is correlated with the performance of the algorithm for that treebank. 


\carlos{The obtained correlations are statistically significant for the sentence lengths that comprise most of the sentences found in actual corpora, both when analyzing projective and non-projective algorithms. In the case of projective algorithms, this factor accounts for more than 25\% of the variance in UAS across algorithms\markda{. This is } a remarkable proportion given the complexity in explaining \markda{how} treebank-specific accuracy 
\markda{differs} between algorithms and the variety of factors involved.}

\carlos{In fact, to the best of our knowledge, this is the first study in which these differences are studied quantitatively between transition-based algorithms of the same search space, thus casting light on a question that has been open since the introduction of the first transition-based parsers in the early 2000s.}

\carlosIII{It is worth noting that the effect of displacement distributions on parsing accuracy is independent of the effect of transition sequence length. It has been hypothesized that short transition sequences reduce error propagation \cite{FerGomNAACL2018b}, but this effect does not help when comparing the relative performance of algorithms on different treebanks, as we do here: for example, the arc-standard and arc-eager algorithms produce transition sequences of identical lengths, independently on the syntactic structures found (they always need exactly $2n$ transitions for a sentence of length $n$). However, as we have seen, their inherent displacement distributions are different and can be used to explain their suitability to different treebanks.}


The insights provided in this paper could be useful to guide parsing algorithm design: since algorithms tend to be more accurate on corpora that are 
close\markda{r} to their inherent distribution, a potential avenue for designing better transition-based parsing algorithms is to try to make their inherent distribution match that of human languages more closely.

\john{To further validate the hypothesis investigated here, it would be interesting to generate artificial treebanks in such a way so as to create a spread of arc distributions so we can control the EMD range.}

\john{Beyond the explicit findings of this paper, it is interesting to observe that linguistic considerations can have an impact in natural language processing systems and more analyses like this, such as considering what makes certain languages harder to model than others, should hopefully prove to be useful in the future \cite{mielke-etal-2019-kind}.}

\section{Conclusion}

\carlos{We have introduced the concept of an algorithm's inherent displacement distribution, which captures the algorithm's bias towards implicitly preferring certain dependency lengths and directions to others. We have shown that given a treebank, the similarity between each transition-based algorithm's inherent \markda{dependency} displacement distribution and the treebank's 
distribution is a strongly correlated to the corresponding algorithm's performance on that treebank.}

\section*{Acknowledgments}
This work has received funding from the European Research Council (ERC), under the European Union's Horizon 2020 research and innovation programme (FASTPARSE, grant agreement No 714150), from the  ANSWER-ASAP project (TIN2017-85160-C2-1-R) from MINECO, and from Xunta de Galicia (ED431B 2017/01, ED431G 2019/01). 
\section{Bibliographical References}
\bibliography{acl2019}
\bibliographystyle{lrec}
\clearpage
\onecolumn
\appendix
\end{document}